\RequirePackage{fix-cm}
\documentclass[twocolumn]{svjour3}          
\smartqed  
\usepackage{graphicx}
\usepackage{graphics}
\usepackage{textcomp}   
\usepackage{colortbl}
\usepackage[table,dvipsnames]{xcolor}

\definecolor{Gray}{gray}{0.85}
\definecolor{SkyBlue}{rgb}{0.88,1,1}

\usepackage{multirow}
\usepackage{boldline}
\usepackage{cellspace}
\usepackage{tabularx}
\usepackage{booktabs}
\usepackage{makecell}
\usepackage{caption}
\usepackage{svg}
\usepackage{float}
\usepackage{tablefootnote}
\usepackage{diagbox}
\usepackage{makecell}

\definecolor{OliveGreen}{rgb}{0,0.6,0}
\definecolor{SoftRed}{rgb}{1,0.2,0.2}

\usepackage{hyperref}
\hypersetup{
     colorlinks   = true,
     citecolor    = OliveGreen,
     linkcolor    = SoftRed
}
\usepackage[hyperpageref]{backref}

\usepackage{caption}
\newcolumntype{M}[1]{>{\centering\arraybackslash}m{#1}}

\begin{document}

\title{Quo Vadis, Skeleton Action Recognition?
}


\author{Pranay Gupta         \and
        Anirudh Thatipelli   \and
        Aditya Aggarwal      \and
        Shubh Maheshwari     \and
        Neel Trivedi         \and
        Sourav Das           \and
        Ravi Kiran Sarvadevabhatla
}


\institute{Ravi Kiran Sarvadevabhatla \at
              F23, 3rd Floor, KCIS, IIIT Hyderabad, Gachibowli, Hyderabad 50032, INDIA \\
              \email{ravi.kiran@iiit.ac.in}           
}

\date{Received: date / Accepted: date}

\maketitle

\begin{abstract}
In this paper, we study current and upcoming frontiers across the landscape of skeleton-based human action recognition. To study skeleton-action recognition in the wild, we introduce Skeletics-152, a curated and 3-D pose-annotated subset of RGB videos sourced from Kinetics-700, a large-scale action dataset. We extend our study to include out-of-context actions by introducing Skeleton-Mimetics, a dataset derived from the recently introduced Mimetics dataset. We also introduce Metaphorics, a dataset with caption-style annotated YouTube videos of the popular social game Dumb Charades and interpretative dance performances. We benchmark state-of-the-art models on the NTU-120 dataset and provide multi-layered  assessment of the results. The results from benchmarking the top performers of NTU-120 on the newly introduced datasets reveal the challenges and domain gap induced by actions in the wild.  Overall, our work characterizes the strengths and limitations of existing approaches and datasets. Via the introduced datasets, our work enables new frontiers for human action recognition. 
\keywords{human action recognition \and human activity recognition \and skeleton \and 3-D human pose \and deep learning}
\end{abstract}

\section{Introduction}
\label{sec:introduction}

Understanding human actions, especially from their 2-D and 3-D joint-based skeleton representations, has received a lot of focus recently. Joint-based representations have a small memory footprint which improves feasibility of on-board processing in compute-restricted environments (e.g. smartphones, cameras on IoT devices). The privacy-friendly nature of the skeleton representation is also an advantageous factor.

On the flip side, obtaining accurate 3-D skeleton data usually requires specialized capture mechanisms and constraints on the capture environment. Even after the capture hurdle is crossed, the sparsity of skeleton representation relative to denser counterparts (RGB, depth) induces ambiguity and imposes additional challenges. In addition, the lack of large-scale, diverse datasets remained a challenge until the advent of datasets such as NTU-60~\cite{Shahroudy_2016_CVPR} and PKU-MMD~\cite{liu2017pku}. These datasets have prompted a number of diverse approaches for skeleton-based action recognition~\cite{Shi_2019_CVPR,Wu_2019_ICCV,peng2020learning,zhang2019view,2sagcn2019cvpr,song2019richly,Li_2019_CVPR}. The introduction of the even larger NTU-120 dataset~\cite{Liu_2019_NTURGBD120} is poised to continue this trend.

The datasets and capture methods for aforementioned works are confined to controlled, indoor settings. Naturally, this prompts the question regarding the ability to recognize human activities occurring outdoors, `in the wild'? Also, in recent times, a number of works on robust estimation of human 3-D pose from RGB data have emerged~\cite{kocabas2019vibe,kolotouros2019spin,RogezWS18}. These prompt yet another question: How well can human actions be recognized in terms of 3-D skeletal pose estimated from RGB videos? To answer these questions, we introduce Skeletics-152 (Sec.~\ref{sec:skeleticsintro}), a carefully curated and 3-D pose-annotated subset of videos sourced from Kinetics-700~\cite{DBLP:journals/corr/abs-1907-06987}, a large-scale RGB action dataset. 

Actions in NTU-120 and Kinetics datasets retain either full or partial context supplied by object interactions and background. In contrast, out-of-context actions represent an unconventional and challenging frontier for skeleton action recognition. To benchmark performance for such actions, we introduce the skeletal version of Mimetics~\cite{weinzaepfel2019mimetics}, a subset of Kinetics-400 containing exaggerated, out-of-context human actions (Sec.~\ref{sec:mimeticsintro}). Additionally, we introduce Metaphorics, a new video dataset with detailed action phrase annotations for videos of the popular social game Dumb Charades and expert  dance performances of popular songs (Sec.~\ref{sec:Metaphoricsintro}).

Typically, the introduction of a newer, larger dataset (NTU-120) is marked by a flurry of novel architectures which aim to solve challenging domain tasks. In this paper, we argue that this is also a good opportunity to evaluate approaches originally trained for earlier dataset versions and more generally, re-evaluate the status quo. This argument has already been made successfully for RGB action recognition~\cite{sigurdsson2017actions}. To this end, we benchmark state-of-the-art approaches on the NTU-120 dataset and analyze the results (Sec.~\ref{sec:ntu120}). Subsequently, we evaluate the performance of top ranked models on our newly introduced datasets -- Skeletics-152 (Sec.~\ref{sec:skeleticsres}), Skeleton-Mimetics (Sec.~\ref{sec:mimeticsres}), Metaphorics (Sec.~\ref{sec:MetaphoricsEvaluation}). 
 
 \begin{figure*}[!ht]
    \centering
    \includegraphics[width=\textwidth]{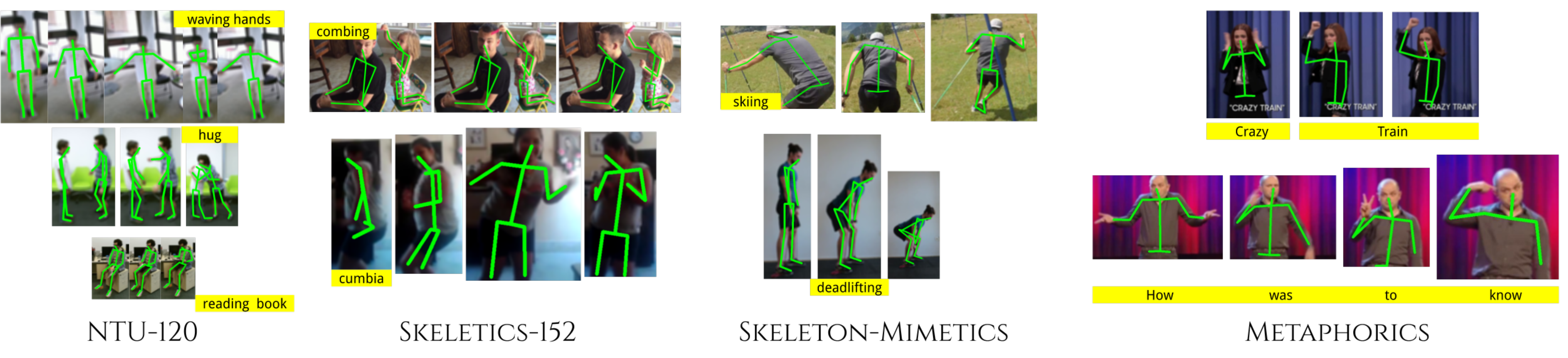}
    \caption{A pictorial illustration of the landscape for skeleton-based action recognition. Datasets such as \textsc{NTU-120} characterize actions in controlled lab-like settings. We use state-of-the-art RGB 3-D pose estimation to obtain skeletons and benchmark recognition models `in the wild' by introducing \textsc{Skeletics-152} dataset (Sec.~\ref{sec:skeleticsintro}). To explore out-of-context action recognition in the wild, we introduce \textsc{Skeleton-Mimetics} (Sec.~\ref{sec:mimeticsintro}) and benchmark models trained on \textsc{Skeletics-152}. As a novel frontier for action recognition, we introduce \textsc{Metaphorics} (Sec.~\ref{sec:mimeticsintro}) which contains indirectly conveyed metaphor-style actions. Note that all datasets are skeleton-based -- RGB background has been included to convey the original context.}
    \label{fig:qv-overview}
\end{figure*}

Overall, our work characterizes the strengths and limitations of existing approaches and datasets. It also provides an assessment of top-performing approaches across a spectrum of activity settings and via the introduced datasets, proposes new frontiers for human action recognition.

Our primary contributions can be summarized as follows:

\begin{itemize}
    \item We introduce Skeletics-152, a curated 3-D pose annotated subset of Kinetics-700 for benchmarking skeleton action recognition `in the wild'(Sec.~\ref{sec:skeleticsintro}).
    \item We introduce Skeleton-Mimetics to recognize skeleton-based out-of-context and exaggerated actions  (Sec.~\ref{sec:mimeticsintro}).
    \item We introduce Metaphorics, a new video dataset with phrase annotations for YouTube videos of Dumb Charades and interpretative dance to explore the new frontier of metaphor-style actions (Sec.~\ref{sec:Metaphoricsintro}).
    \item We benchmark current, past state-of-the-art skeleton action recognition approaches on in-lab datasets (Sec.~\ref{sec:ntu120}) and also on our newly introduced datasets containing actions happening in-the-wild (Sec.~\ref{sec:skwild}). We also summarize trends within and across these various datasets (Sec.~\ref{sec:qualitative}).
\end{itemize}

To enable a rich, interactive exploration of our contributions mentioned above, we have made them available at \url{https://skeleton.iiit.ac.in/}. The website features an interactive data analytics dashboard, code and pre-trained models for top-performing skeleton action recognition models and
new skeleton action datasets (Skeletics-152, Skeleton-Mimetics, Metaphorics) introduced by us for additional exploration and benefit of the community.

\section{Related Work}

\noindent \textbf{Skeletal Datasets:} Numerous 3-D skeletal datasets ~\cite{cg-2007-2,Liu_2019_NTURGBD120,Shahroudy_2016_CVPR,liu2017pku,seidenari2013recognizing,li2010action} have been proposed over the last decade to further the advances in human action understanding. These datasets are focused on sequence based action detection and involve human subjects performing daily actions captured from multiple viewpoints. Sadeghipour et. al.~\cite{3DICONIC} introduce 3D Iconic gesture dataset where subjects outline the shape of virtual objects that is captured via Kinect v2 sensor. Recent work by Yan et al.~\cite{stgcn2018aaai} introduces Skeleton-Kinetics using 2-D Open Pose on the large scale video dataset Kinetics-400. Weinzaepfel et. al. take this idea further and introduce Mimetics~\cite{weinzaepfel2019mimetics} containing a subset of Kinetics-400 with mimed actions. 

\noindent \textbf{Skeleton Action Recognition:} An earlier era of works serve to document handcrafted features for skeleton action recognition~\cite{Vemulapalli2014HumanAR,wang2012mining,covdesc,halim2016human}. The recent class of approaches based on deep networks can be broadly categorized into three groups based on input skeleton data representation.

The first group explicitly consider the sequential nature of actions wherein the temporal dependencies are modelled using an RNN or an LSTM~\cite{zhang2017view,DBLP:journals/corr/KimR17,DBLP:journals/corr/ZhuLXZLSX16}. To further discriminate activities based on the joint dependencies, Song et al.~\cite{DBLP:journals/corr/SongLXZL16} introduce attention mechanisms at multiple levels in the network. Kundu et. al.~\cite{DBLP:journals/corr/abs-1812-02592} learn the action sequence as a trajectory in the pose manifold for the downstream activity classification task. Caetano et al.~\cite{Caetano:AVSS:2019} use CNN-based feature representation over a temporal window containing skeleton dynamics.

The second group of works model the input skeleton as a single spatio-temporal unit. In some instances, this unit is a tensor of the form $\mathsf{frames} \times \mathsf{joints} \times \mathsf{coordinates}$ which is subsequently processed by a CNN~\cite{DBLP:journals/corr/LiZXP17,DBLP:journals/corr/LiDCCLH17,7486569,Caetano:SIBGRAPI:2019,HCN}. More recently, a series of approaches use graph convolutions to model the (spatio-temporal) unit. Prominent examples include the ST-GCN framework introduced by Yan et. al.~\cite{stgcn2018aaai} and variants~\cite{Wu_2019_ICCV,song2019richly}. In contrast to the fixed graph in ST-GCN, newer approaches involve adaptation to learn graph topology~\cite{Shi_2019_CVPR,peng2020learning,2sagcn2019cvpr,Si_2019_CVPR,Tang_2018_CVPR}.

In addition to the groups mentioned above, hybrid approaches also exist. Si et al.~\cite{Si_2019_CVPR} employ an attention-based graph convolutional LSTM to capture the spatio-temporal co-occurrence relationships. Zhang et. al.~\cite{zhang2019view} propose a CNN-RNN late-fusion model with learnable view transformation. For a survey of 3-D skeleton action recognition, refer to Presti et al.~\cite{presti20163d} and Wang et. al.~\cite{WANG2018118}.

\noindent \textbf{Skeleton Action Recognition from RGB video based pose:} In another class of approaches, human skeletal pose estimated from in-the-wild RGB video frames is used for action recognition. A number of  approaches based on 2-D skeleton pose from RGB video exist~\cite{angelini2018actionxpose,eweiwi2014efficient,ayumi2016pose,cheron2015p,Jhuang_2013_ICCV}. A recent variation involves a pseudo 3-D pose representation wherein 2-D OpenPose coordinates~\cite{cao2018openpose} in Kinetics-400~\cite{Carreira2017QuoVA} videos are augmented with joint-level confidence scores as the third coordinate~\cite{Shi_2019_CVPR,peng2020learning,2sagcn2019cvpr,Li_2019_CVPR,stgcn2018aaai}.
\begin{figure*}[!ht]
\centering
    \includegraphics[width=\textwidth]{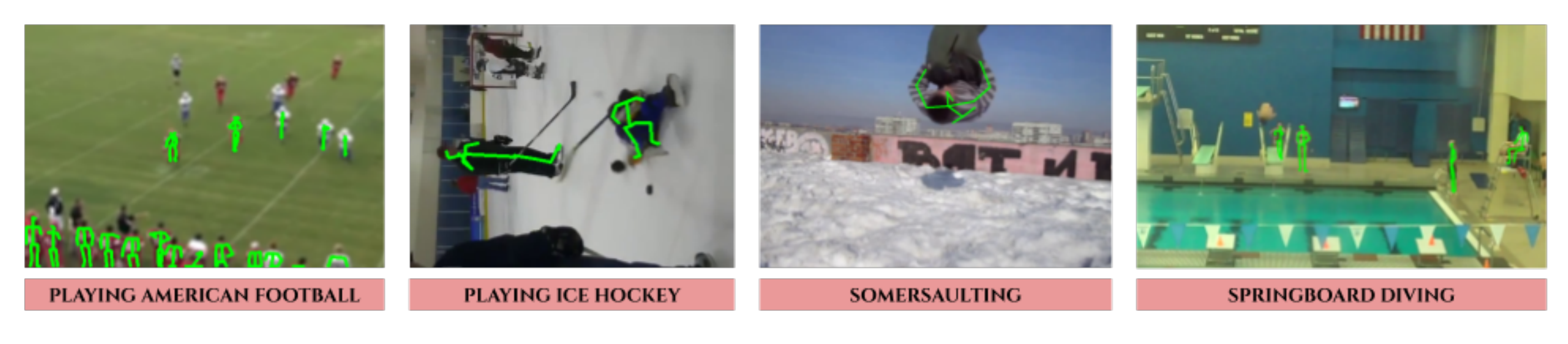}
\caption{Examples of classes from  Kinetics-700 omitted for skeleton action recognition. In `Playing American football', multiple people are detected. For `Playing ice hockey'   and `Somersaulting', pose estimation is not accurate. In `Springboard diving', the person performing the diving action is not tracked.}
\label{fig:vibe_examples}
\end{figure*}

\section{Datasets}
\label{sec:nov_skel_wild}

As mentioned in the Introduction (Section~\ref{sec:introduction}), large-scale datasets such as NTU-120 represent lab-style, controlled, indoor settings. In contrast, a much larger variety of human actions characterize in-the-wild RGB videos of human activities. To obtain 3-D skeleton representations from such videos, pose estimation techniques are applied on action sequences from large-scale activity datasets. In this section, we explore three diverse settings with progressively increasing level of complexity and abstractness in terms of human actions.

\subsection{Skeletics-152}
\label{sec:skeleticsintro}

To source in the wild action videos, we use Kinetics-700~\cite{DBLP:journals/corr/abs-1907-06987} as the starting point. Kinetics-700 is a large-scale video dataset consisting of over $650{,}000$ YouTube video clips spanning over $700$ action categories ranging from daily routine activities, sports and other fine-grained actions. However, unlike previously existing dataset with similar preparatory approach (Skeleton-Kinetics-400~\cite{stgcn2018aaai}), we carefully omit categories from action settings which are incompatible for pose-based skeleton action recognition. Specifically,
 
\begin{itemize}
\item A number of classes (e.g. `Petting cat', `Scrubbing face') were removed because most of the videos contain occluded poses which make the 3D pose estimation unviable.
\item Some classes (e.g. `Cooking eggs', `Wrapping presents', `Clay pottery making') were removed as they were captured from egocentric views.
\item Some classes (e.g. `Peeling apples', `Peeling potatoes', `Baking cookies') are highly object-centric and hence, irrelevant for skeleton based action recognition.
\item Classes involving no substantial movement (e.g. `Staring', `Attending a conference') cannot be recognised solely based on human pose.
\item Classes where the labels differ solely due to scene background were removed (e.g. `Walking through snow'   is same as `Walking').
\end{itemize}

\begin{figure*}[!t]
    \centering
    \includegraphics[width=\textwidth]{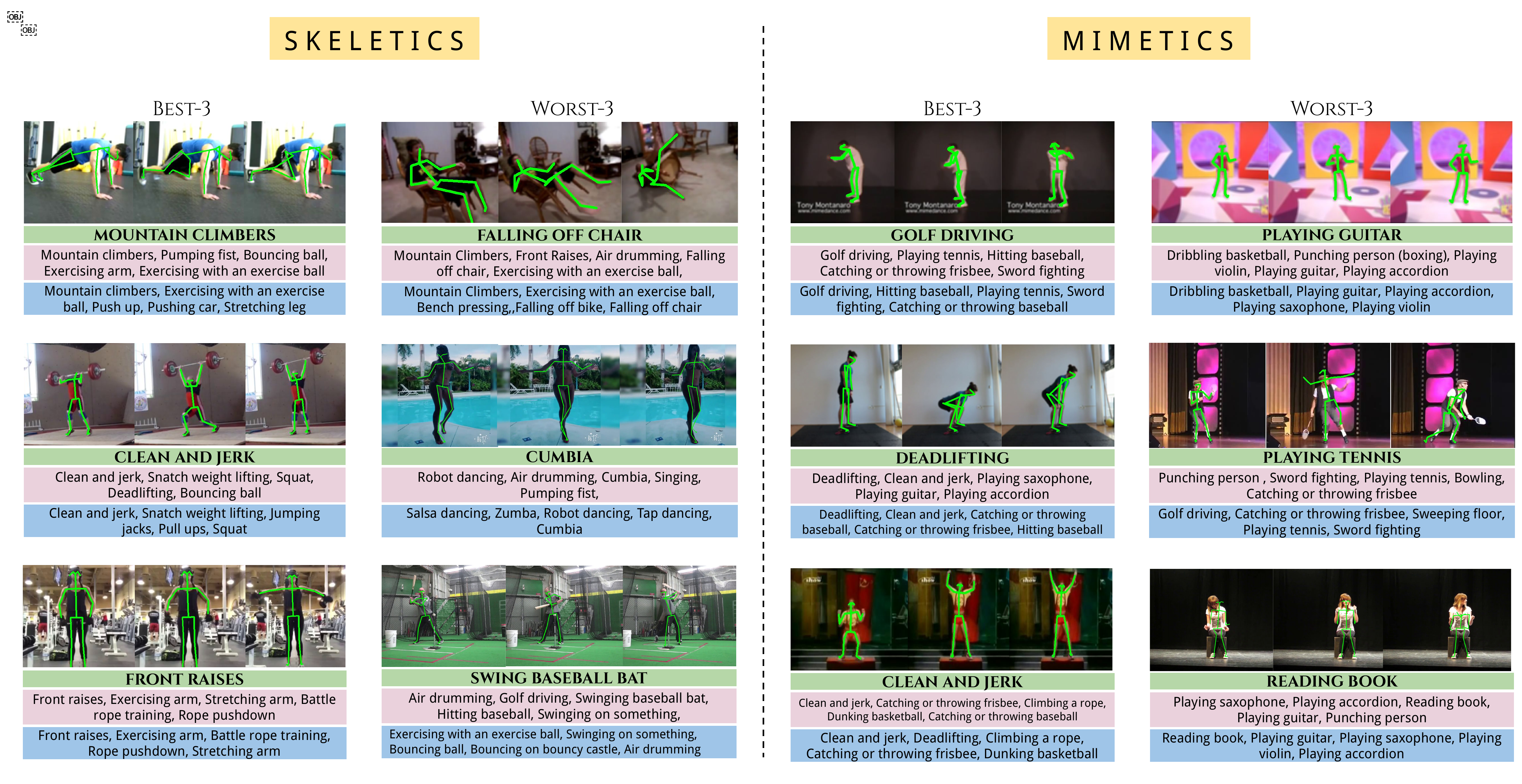}
    \caption{Sample skeleton sequences from Skeletics-152 and Mimetics-Skeleton. The sequences are chosen from best-3 and worst-3 classes in terms of performance achieved by best models on these datasets (see Tables~\ref{tab:Kinetics-results},~\ref{tab:Mimetics-test-results}). The ground-truth phrase is color-coded green. The top-5 predictions by 4s-ShiftGCN  are coded pink and those by MS-G3D are coded blue. Refer to Section \ref{sec:skwild} for details on the evaluation protocol and predictions.}
    \label{fig:skeletics-mimetics-3}
\end{figure*}

After carefully removing such action categories, we use VIBE~\cite{kocabas2019vibe} on the remaining $274$ categories to obtain the corresponding 3-D skeleton sequences. Within these sequences, we removed classes such as `Playing American football', `Playing ice hockey', `Doing aerobics' containing large groups of people performing different activities within a single video. In addition, classes such as `Somersaulting', `Springboard diving'  were removed since the VIBE model typically reported missing joints. Figure \ref{fig:vibe_examples} shows some examples of omitted action classes. 

For the case of multiple ($> 2$) skeleton detections in single video, we select the top two skeletons appearing in maximum number of frames. For intermediate frames with missing skeletons, we perform bounding box and joint interpolation. In the end, we obtain Skeletics-152, our curated 3-D skeleton dataset which contains $125{,}621$ sequences spread over $152$ classes. Refer to Figure~\ref{fig:skeletics-mimetics-3} (left) for some example sequences from Skeletics-152.

It must be noted that Skeletics-152 is different from  Skeleton-Kinetics-400~\cite{stgcn2018aaai} which contains pseudo 3-D pose [2-D Pose + Joint-level confidence] obtained from the OpenPose~\cite{cao2018openpose} toolbox. Skeleton-Kinetics-400 indiscriminately includes all categories, without any curation. Unlike VIBE-based skeletons in our dataset, the pseudo 3D Skeleton-Kinetics-400 representations fail to capture the actual dynamics of 3D motion. 

\begin{figure*}[!ht]
\centering
\includegraphics[width=\textwidth]{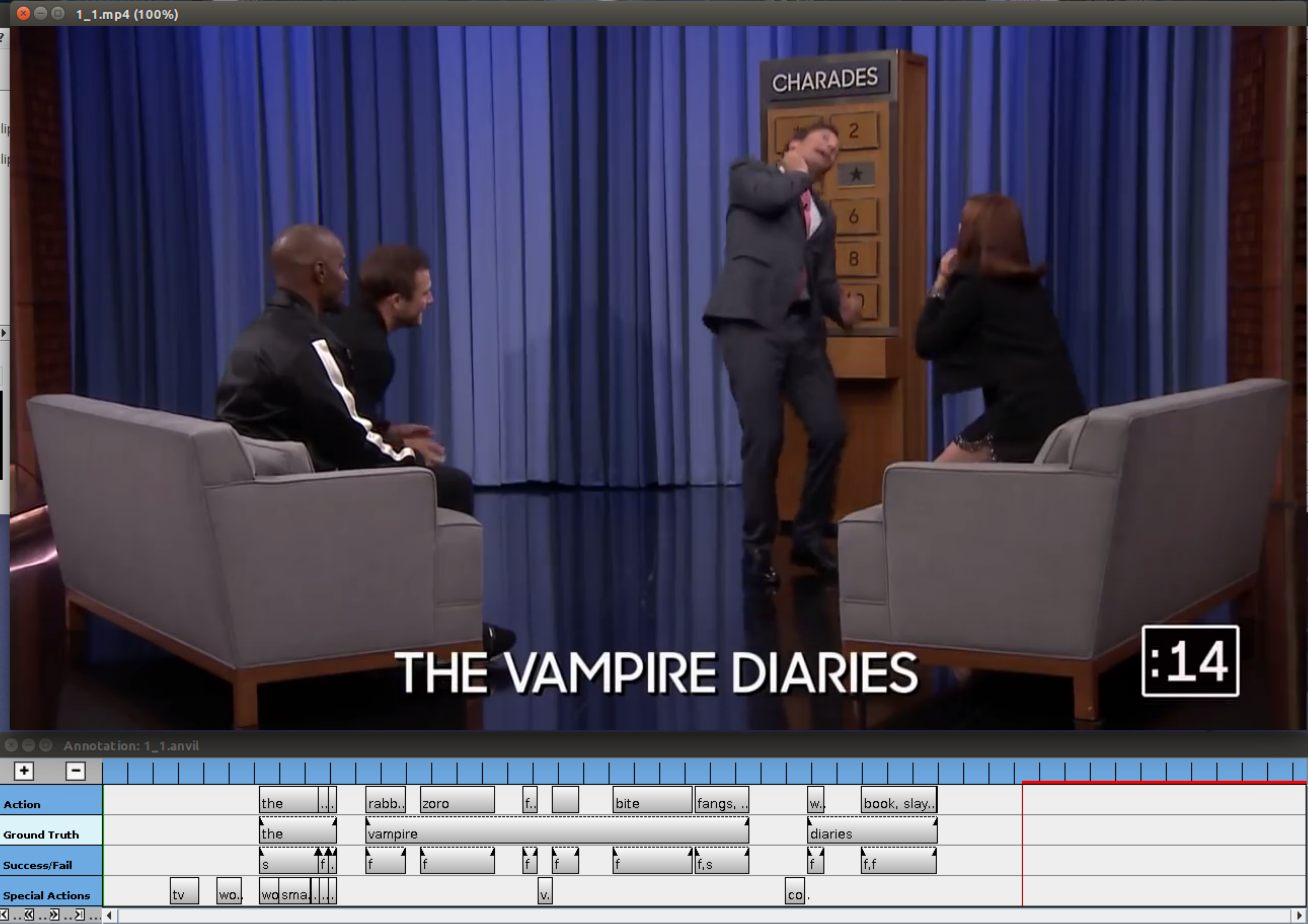}
\caption{An illustration of the annotation for a typical Charades episode using the Anvil interface. `Action', `Ground Truth', `Success/Fail', `Special Actions' are the annotation channels. In the `Action'channel, `rabb...(rabbit)'and `zorro' are guesses that the guessing player makes for the first two actions performed by the actor performs upon being revealed the ground truth  phrase `the vampire diaries'. The segment labelled `vampire'  in the `Ground Truth' channel is the entire duration for which the actor tried to act out the word `vampire'. The `Success/Fail'  channel shows the success and failure for corresponding guesses present in the `Action' channel. Here, `rabbit' and `zorro' are both incorrect and hence they are marked as `F'. The `Special Action' channel has tabs containing `TV' and `wo...(number of words)'. These are helping actions to indicate that the phrase is the name of a TV show and the number of words in the phrase respectively.}
\label{fig:anvil}
\end{figure*}

\subsection{Skeleton-Mimetics}
\label{sec:mimeticsintro}

Human actions in the wild tend to be contextual. Context such as background, presence of certain objects can be very influential in certain traditional (RGB) action recognition approaches. This can lead to incorrect predictions when such actions are performed in out-of-context scenarios~\cite{weinzaepfel2019mimetics}. To this end, Mimetics dataset, consisting of $50$ out-of-context action classes and derived from Kinetics-400 was introduced in~\cite{weinzaepfel2019mimetics}. To explore  skeleton action recognition for out-of-context, exaggerated action sequences, we introduce Skeleton-Mimetics. This dataset is derived from Skeletics-152, introduced previously. To create this new dataset, we shortlisted classes with exaggerated movements and gestures. Instead of considering all the videos, we select specific action videos where action is performed by mimicry experts in out-of-context settings and without object interactions. The final dataset consists of $319$ skeleton sequences across $23$ classes.

\begin{figure*}[!ht]
    \centering
    \includegraphics[width=0.8\textwidth]{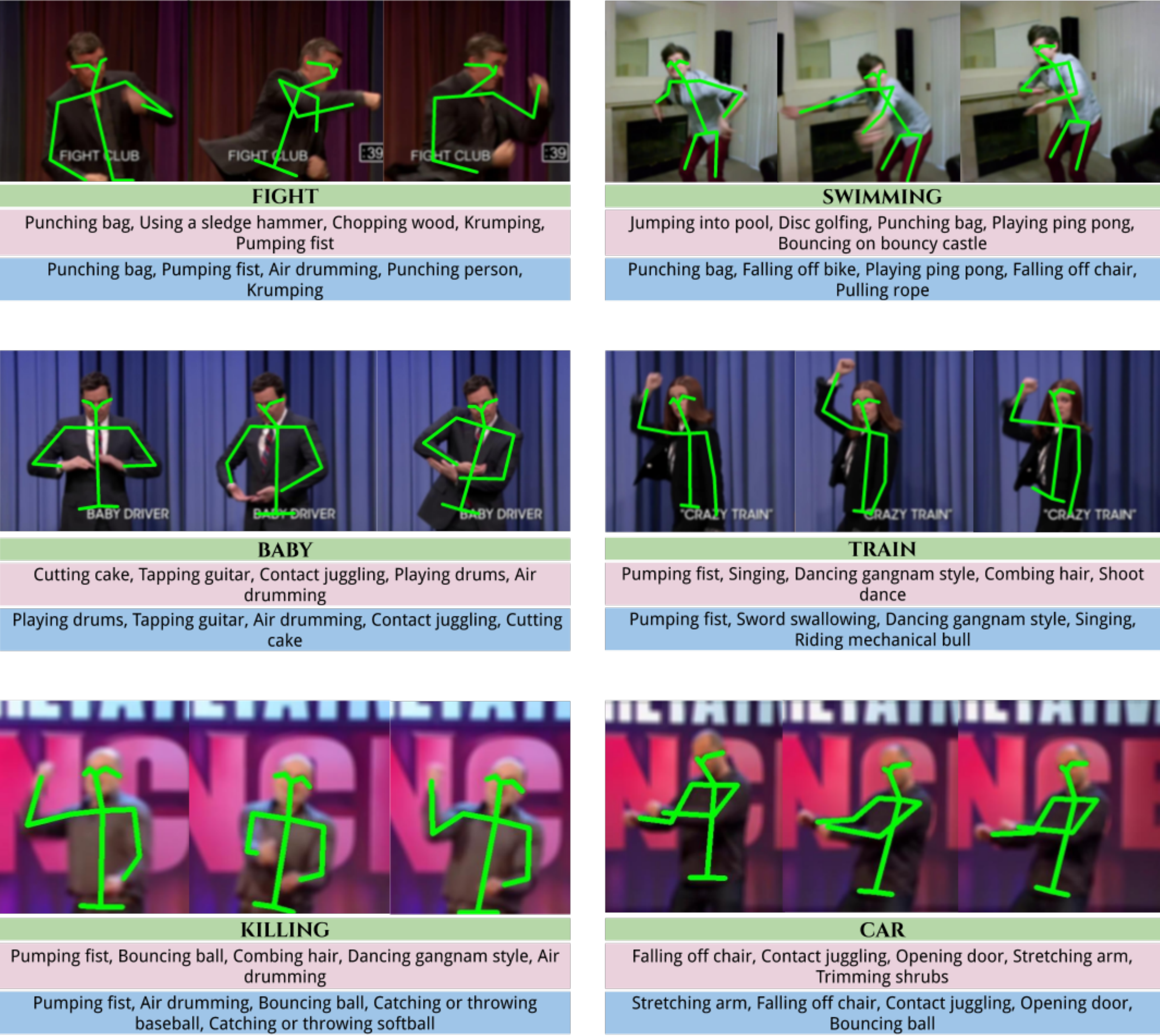}
    \caption{Sample skeleton sequences from our Metaphorics dataset. The ground-truth phrase is color-coded green. The top-5 predictions by 4s-ShiftGCN are coded pink and those by MS-G3D are color-coded blue. Refer to Section \ref{sec:MetaphoricsEvaluation} for details on the evaluation protocol and predictions.}
    \label{fig:metaphorics-results}
\end{figure*}

The proposed dataset differs from the existing Mimetics dataset in terms of accurate 3-D poses. Since Skeleton-Mimetics is derived from a curated set of videos specially designed for skeleton action recognition, it eliminates the factor of unusable 3-D poses for action classification. Further, since we wanted to compare the efficacy of Skeletics-152 as a dataset for pre training skeleton action recognition models, we keep joint positions same for both the datasets. 
Refer to Figure~\ref{fig:skeletics-mimetics-3} (right) for some example sequences from Skeleton-Mimetics.

\subsection{Metaphorics}
\label{sec:Metaphoricsintro}

The datasets encountered so far can be characterized as \textit{verb}-based actions, since the action class is fundamentally incomplete without the verb or the activity being performed. However, humans also tend to associate actions to non-verb words or objects, Iconic Gestures dataset~\cite{3DICONIC} being a known example. In general, actions can be more abstract and used to convey metaphorical concepts. One such scenario is the popular social game of Dumb Charades. The game involves interactive and adaptive guessing of a target `phrase`(usually a movie title) based on actions being performed by an `actor'. Unlike other datasets, nouns and adjectives can have action depictions. Moreover, the vocabulary is open-ended, further compounding the action understanding challenge. To study actions arising in this challenging scenario, we introduce Metaphorics, an even complex dataset. The dataset contains videos from two scenarios - dumb charades and interpretive dance.

\begin{table*}[!ht]
\resizebox{\textwidth}{!}
{%
  \centering
    \begin{tabular}{c|cccccccc}
       \toprule
        & \textsc{\textbf{\thead{No of.\\ Classes}}} & \textsc{\textbf{{\thead{No. of.\\ sequences}}}} & \textsc{\textbf{{\thead{Action\\ Vocabulary}}}} & \textsc{\textbf{\thead{Action\\ setting}}} & \textsc{\textbf{\thead{Action\\ environment}}} & \textsc{\textbf{\thead{Typical\\ action\\ duration}}} & \textsc{\textbf{\thead{Level of\\ action\\ explicitness}}} & \textsc{\textbf{\thead{Camera}}} \\
        \midrule
        
        HDM05 \cite{cg-2007-2} & $100$ & $1500$ & Fixed & Lab/Prompted & Non-contextual & Not specified & High & Fixed \\
        
        3D-Iconic dataset \cite{3DICONIC} & $20$ & $1739$ & Fixed & Lab/Prompted & Non-contextual & Not specified & High & Fixed \\
        
        Florence-3D \cite{seidenari2013recognizing} & $9$ & $215$ & Fixed & Lab/Prompted & Non-contextual & Not specified & High & Fixed \\
        
        NTU-60 \cite{Shahroudy_2016_CVPR} & $60$ & $56880$ & Fixed & Lab/Prompted & Non-contextual & 1-10 seconds & High & Fixed \\
        
        Large-RGB+D \cite{large2016rgb} & $94$ & $4953$ & Fixed & Lab/Prompted & Non-contextual & Not specified & High & Fixed/Moving \\

        Kinetics-skeleton \cite{stgcn2018aaai} & $400$ & $300{,}000$ & Fixed & Wild & Contextual & 10 seconds &  High & Fixed/Moving \\
        
        NTU-120 \cite{Liu_2019_NTURGBD120} & $120$ & $114{,}480$ & Fixed & Lab/Prompted & Non-contextual & 1-10 seconds & High & Fixed \\
        
        Mimetics \cite{mimetics} & $50$ & $713$ & Fixed & Wild & Non-contextual & 1-10 seconds & Moderate & Fixed \\
        
        \rowcolor{Gray}
        Skeletics-152 & $152$  & $125{,}657$ & Fixed & Wild & Contextual & 10 seconds & High & Fixed/Moving \\

        \rowcolor{Gray}
        Skeleton-Mimetics & $23$ & $319$ & Fixed & Wild & Non-contextual & 1-10 seconds & Moderate & Fixed \\
        
        \rowcolor{Gray}
        Metaphorics & N.A & $845$ & Open-ended & Wild/Prompted & Non-contextual & 1-3 seconds & Low & Fixed \\
        
         \bottomrule
    \end{tabular}
    }
    \captionof{table}{Attributes of different datasets in the skeleton based action recognition domain. Gray-shaded rows correspond to the new datasets introduced in this paper. Prompted means that subjects were instructed what action to perform. N.A  means that there is no notion of classes. Not Specified means that the duration of an action is not specified in the respective paper.}
\label{tab:datasets}
\end{table*}

\noindent \textit{Dumb Charades:} We first source Dumb Charade game episodes from YouTube. In the game episodes, one person (`actor') acts out a target phrase word by word while the other player tries to guess the target phrase solely from actions performed by the `actor'. For annotation, we use the popular Anvil tool \cite{kipp2001anvil}.  We annotate (i) target phrase (ii) beginning and ending timestamps for each action segment (iii) guess phrase associated with a segments (iv) episode outcome (`correctly guessed',`incorrect'). We also annotate certain special actions such as number of words and the current word number for a multi-word target phrase. These special actions also include helping actions which the actor uses to convey some basic information to the guessers such as length of the word (`long',`short'). Figure~\ref{fig:anvil} provides an illustration of a typical annotation for a Charades video episode.

To characterize performance of action recognition approaches, we associate each word with its corresponding temporal video segment. After removing instances where the actor is occluded, we obtain $716$ segments across $28$ game sessions.

\noindent \textit{Interpretative dance:} We also source YouTube videos containing interpretative dances of popular songs. In these videos, the song is made audible only to the actor who then proceeds to enact real-time actions corresponding to song lyrics. In this case, the guesser needs to correctly guess the song title based on the performed lyric-based actions. Unlike Charades, the actor is required to act out the song lyrics in real time which increases the challenge since the actions are more fast paced than Charades. 

As part of the annotation process, we align the lyric subtitle file of original song and the video based on the starting point of the song (in the video). Since the actions are  performed in real time, we obtain the action level annotations by aligning the audio file of the video with the original audio file of the song. The temporal extents of the dance are thus annotated into word-level action video segments. We obtain a total of $129$ video segments across two full-length music videos.

\begin{figure}[!t]
    \centering
\includegraphics[width=0.6\linewidth]
 {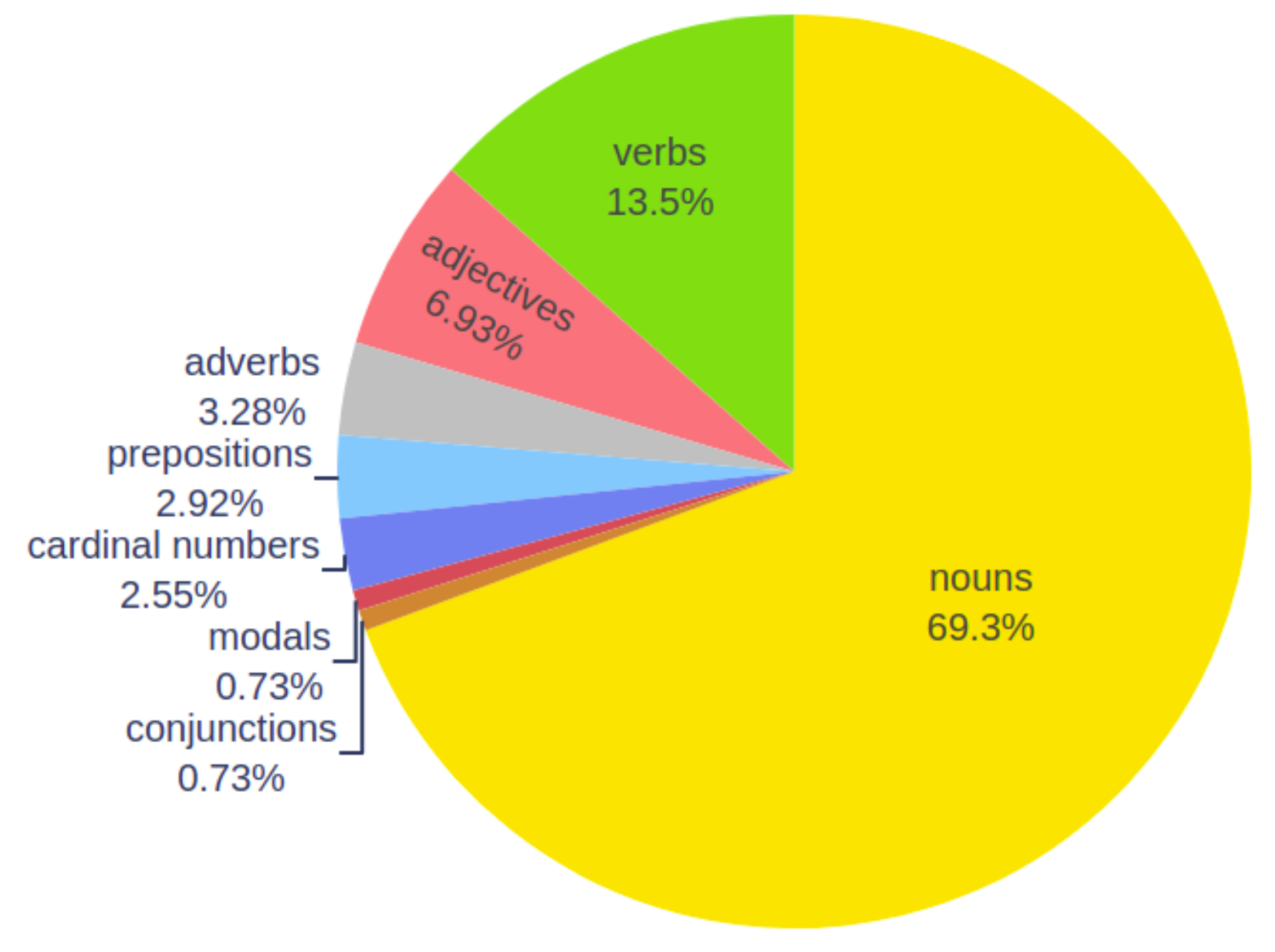}
\caption{Part of speech distribution across the ground-truth for Metaphorics dataset.}
\label{fig:metaphorcs-pos}
\end{figure}

In total, our Metaphorics dataset contains $845$ video clips. As with other RGB datasets, we obtain corresponding 3-D skeleton sequences using VIBE~\cite{kocabas2019vibe} - see Figure~\ref{fig:metaphorics-results} for examples. The proposed Metaphorics dataset is very diverse in terms of the action sequences and the labels due to its open-ended vocabulary (see Figure \ref{fig:metaphorcs-pos}).  Compared to existing datasets, videos tend to be `bursty'  due to the extremely small temporal extents of the actions. The 3D-Iconic dataset~\cite{3DICONIC} is similar to Metaphorics in the sense that it contains object based gesture actions. However, subjects are prompted to explicitly outline shape of the object categories unlike the unprompted actions seen in Metaphorics.

To gain an overall perspective about the datasets, both existing ones and those introduced in our work, we summarize their prominent attributes in Table \ref{tab:datasets}.

\section{Skeleton Action Recognition in the Lab}
\label{sec:ntu120}

In this section, we look at the NTU-120~\cite{Liu_2019_NTURGBD120} dataset which is currently
the largest lab styled 3-D skeleton action recognition dataset. It comprises $114{,}480$ $25$-joint 3-D skeleton annotated videos of $120$ human actions, performed by $106$ subjects in a controlled indoor setting and captured from $32$ different setups.

\subsection{Evaluation Protocol}

Two standard evaluation protocols are typically used for
evaluation of multi-subject multi-viewpoint skeleton action
recognition approaches. In the Cross Subject protocol, the
train and test set are split based on performer id. Under the
protocol proposed by Liu et al.~\cite{Liu_2019_NTURGBD120} for NTU-120, $53$ subject ids out of $106$ are allocated for training and the remaining
for test. We use data from $11$ ($20\%$) randomly selected ids of original training set for validation.

The other protocol is Cross Setup. By default, action
sequences from the $16$ even-numbered camera setup ids are
used for training and $16$ odd setup ids are used for testing.
As with cross subject protocol, we retain the original NTU120 test set and use $4$ ($25\%$) ids randomly chosen from even setup ids for validation.

\subsection{Performance with full sequences}

For benchmarking, we selected approaches which report performance on NTU-120 and top 5 approaches with the best performance on NTU-60~\cite{Shahroudy_2016_CVPR}, the precursor to NTU-120. The results on the test set of NTU-120 can be viewed in Table~\ref{tab:ntu120-val}. The results show that 4s-Shift-GCN~\cite{cheng2020shiftgcn} and MS-G3D~\cite{liu2020disentangling} are the best performers for Cross Setup and Cross Subject respectively.

The gray-shaded portion of Table~\ref{tab:ntu120-val} shows the performance of top performing NTU-60 models evaluated on the NTU-120  test set. Note that these models were not originally designed for NTU-120 and were retrained by us from scratch, for benchmarking purposes. From the results, we notice that our version of VA-NN~\cite{zhang2019view}, retrained with a more powerful backbone (ResNeXt-101) performs competitively with state-of-the-art NTU-120 approaches (MS-G3D and 4s-Shift-GCN). VA-CNN has a relatively simpler architecture and is fast to train , adding to its appeal. More significantly, the results underscore the importance of  benchmarking existing approaches on newly introduced datasets while investing effort into creation of novel architectures.

\begin{table}[!t]
\resizebox{\linewidth}{!}
 {%
  \centering
 \begin{tabular}{l|c|c}
 \toprule
             Method & Cross Setup & Cross Subject\\
 \midrule
 \rowcolor{Gray}
  DGNN~\cite{Shi_2019_CVPR} & $78.13$  & $75.16$  \\
  \rowcolor{Gray}
  GCN-NAS~\cite{peng2020learning} & $85.29$ & $81.99$ \\
  2s-SDGCN~\cite{Wu_2019_ICCV} & $86.18$ & $84.42$ \\
  VA-CNN (ResNeXt-101)~\cite{zhang2019view}\tablefootnote{Our version of VA-CNN~\cite{zhang2017view} with ResNeXt-101 backbone.} & $86.90$ & $84.88$ \\
  4s-ShiftGCN~\cite{cheng2020shiftgcn} & $\mathbf{87.65}$ & $85.76$ \\
  MS-G3D~\cite{liu2020disentangling} & $87.32$ & $\mathbf{85.92}$\\
  \midrule
  best-5 models average pooled  & $\mathbf{88.80}$  & $\mathbf{87.22}$\\
  \bottomrule
 \end{tabular}
  }
\caption{\label{tab:ntu120-val} Benchmarking comparison for NTU-120 test set (mean accuracy). Gray-shaded lines correspond to models which originally reported results on NTU-60 but retrained by us on NTU-120 for comparison. The last row of the table corresponds to the new state-of-the-art which is average pooled ensemble of best-5 models above.}
\end{table}

\begin{figure}[!t]
\centering
    \includegraphics[width=\linewidth]{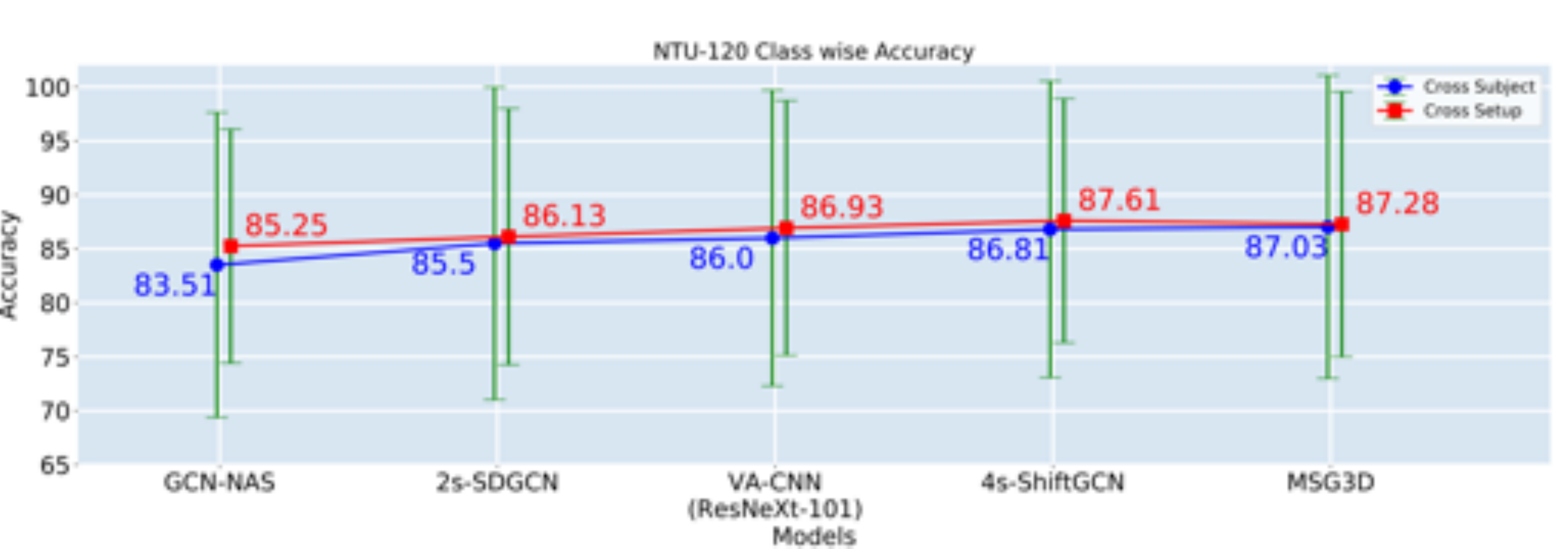}
\caption{Class accuracy plots for NTU-120 with standard deviation}
\label{fig:ntu120-class}
\end{figure}

Figure~\ref{fig:ntu120-class} shows the mean accuracy and associated standard deviation for the top-5 models. The significant magnitude of deviation indicates that additional progress is needed before mean accuracy can be considered a reliable measure of overall performance.

\begin{figure*}[htbp]
\centering
    \includegraphics[width=0.98\linewidth]{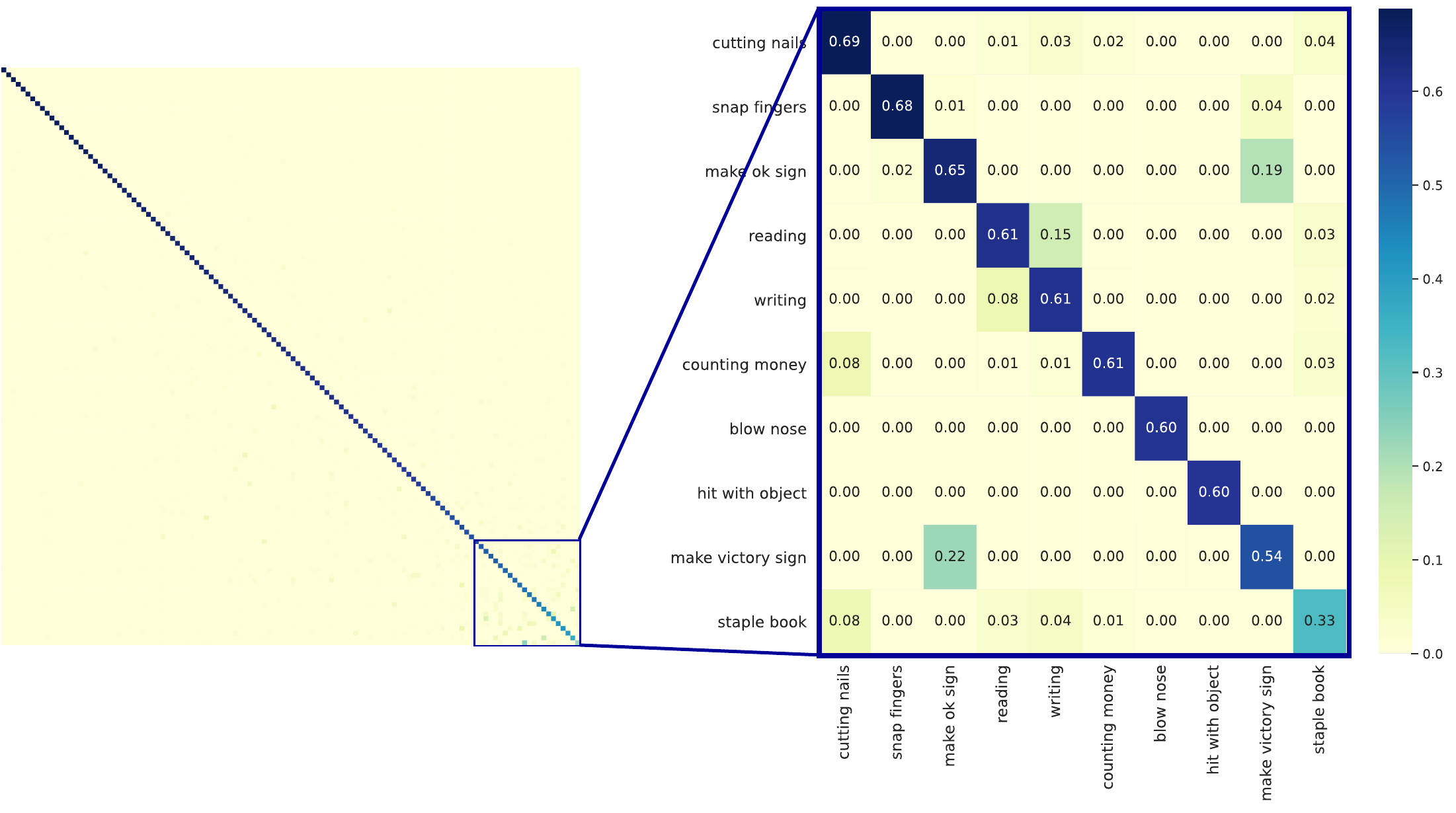}
\caption{The NTU-120 confusion matrix for MS-G3D model sorted by class-wise accuracy shows that the least accurately recognized classes are confused amongst each other (magnified inset).}
\label{fig:ntu-120_confusion}
\end{figure*}

To obtain a better understanding of performance, we list the 10 best recognized and worst recognized action classes in Table \ref{tab:top-bottom-NTU-csub} (cross subject) and Table  \ref{tab:top-bottom-NTU-cset} (cross setup). The results show that the best and worst performers largely stay same across all the models. The best performing classes (e.g. `Arm swings`, `Jump up`, `Walking towards') have distinct actions and involving large joint-level movements. On the other hand,  action classes containing subtle actions with fine-grained differences are hardest to recognize and exhibit large intra-class confusion (Figure \ref{fig:ntu-120_confusion}). For instance, `Make ok sign' and `Make victory sign'    get confused with each other significantly because the actions differ only in terms of hand joint movement which is not captured in adequate detail  by the Kinect sensor. In addition, skeletons for classes such as `Reading'`, `Writing'  have very low inter-class variability, resulting in poor performance.

The top 5 models  exhibit similarities at the set level for best-10 and worst-10 classes. However, motivated by the variance in actual rank order (Tables \ref{tab:top-bottom-NTU-csub},\ref{tab:top-bottom-NTU-cset}), we examine performance with an average pooled ensemble of top-5 models. The noticeably improved ensemble performance  (bottom row of Table~\ref{tab:ntu120-val}) suggests that the top-5 models span action classes in a complementary manner.

\begin{figure}[!t]
\centering
    \includegraphics[width=\linewidth]{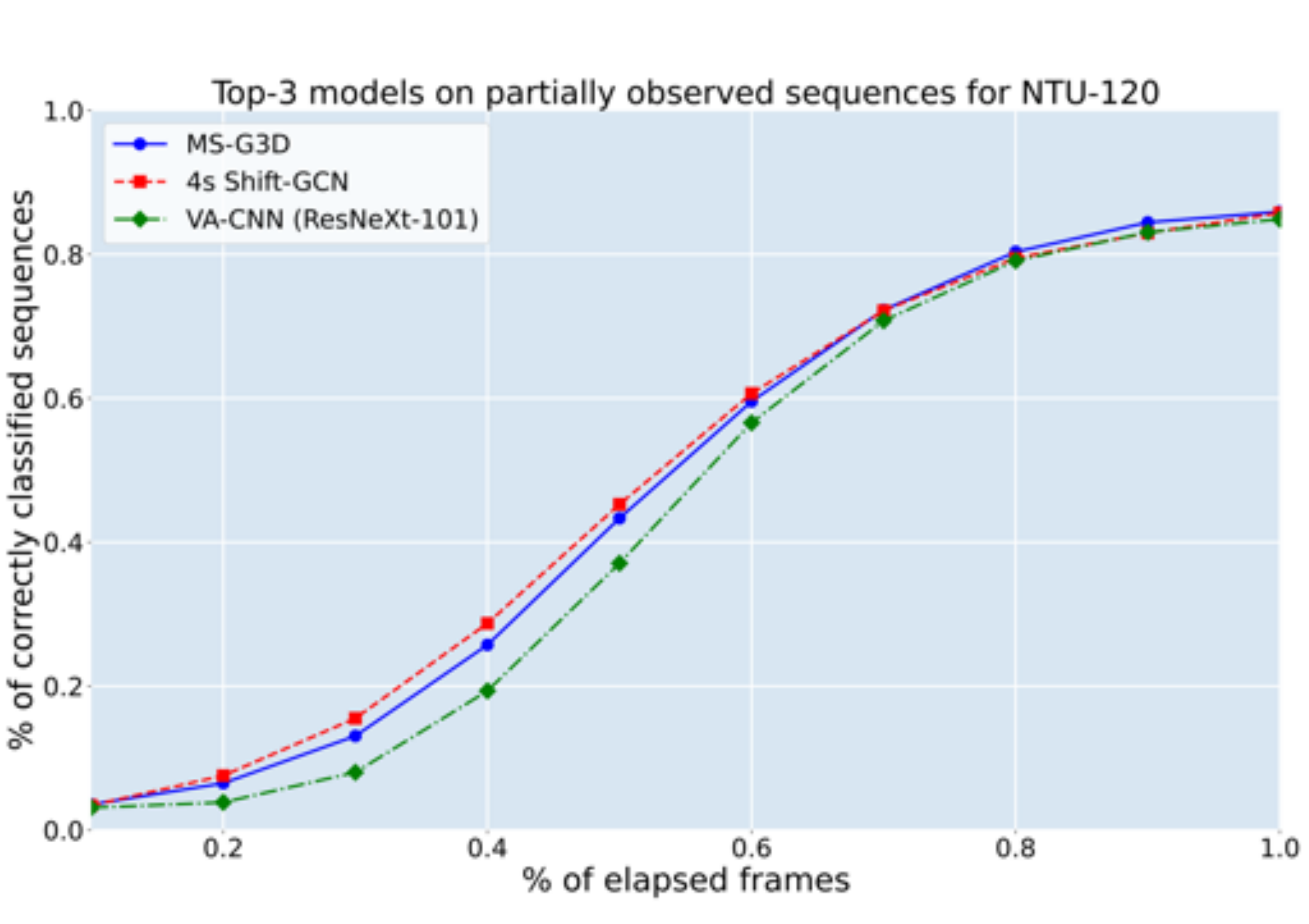}
\caption{Comparison of top-3 models on the partially observed sequences for NTU-120 (Cross Subject)}
\label{fig:ntu120-temporal}
\end{figure}

 \begin{table*}[!]
 \centering
 \resizebox{\textwidth}{!}{%
 \begin{tabular}{l|ccccc}
  \toprule
             & \textbf{MS-G3D} & \textbf{4s-ShiftGCN} & \textbf{VA-CNN} & \textbf{2s-SDGCN} & \textbf{2s-AGCN} \\
  \midrule
    \multirow{10}{4.5em}{\centering Best-10}
    & Hugging  & Jump up & Falling down & Stand up & Hugging  \\ 
    & Hopping  & Staggering & Walking towards & Jump up & Drink after cheers\\ 
    & Put on jacket & Take off jacket & Jump up & Walking towards & Arm swings \\
    
    & Walking Towards & Arm swings & Arm swings & Drink after cheers & Put on jacket \\
    
    & Drink after cheers & Put on jacket & Pushing & Arm circles & Run on the spot\\ 
    & Jump up & Walking towards & Staggering & Put on Jacket & Arm circles \\
    
    & Staggering & Arm circles & Arm circles & Arm swings & Jump up \\
    
    & Arm circles & Hugging & Squat down & Staggering & Staggering \\ 
    & Arm swings & Run on the spot & Drink after cheers & Take off jacket & Walking towards\\
    
    & Capitulate & Drink after cheers & Follow & Hugging & Follow\\ 
  \midrule
    \multirow{10}{5em}{\centering Worst-10}
    &  Staple book & Staple book & Make ok sign & Staple book & Staple book  \\ 
    &  Make victory sign & Make victory sign & Make victory sign & Make victory sign & Make ok sign  \\ 
    & Hit with object & Make ok sign & Staple book & Writing & Make victory sign  \\ 
    & Blow nose & Counting money & Counting money  & Counting money & Counting money \\ 
    & Counting money & Blow nose & Play with phone or tablet & Make ok sign & Writing \\ 
    & Writing  & Reading & Reading & Cutting nails & Cutting paper \\ 
    & Reading  & Cutting paper & Writing & Blow nose & Hit with object  \\ 
    & Make ok sign  & Hit with object & Hit with object & Play with phone or tablet & Blow nose  \\ 
    & Snap fingers  & Cutting nails & Fold paper & Wield Knife & Cutting nails\\ 
    & Cutting nails & Play with phone or tablet & Play magic cube & Hit with object & Yawn  \\ 
  \bottomrule
  \end{tabular}
  }
  \captionof{table}{Best-10 and Worst-10 classes for models trained on NTU-120 (Cross Subject) }
  \label{tab:top-bottom-NTU-csub}
\end{table*}
 \begin{table*}[!]
 \centering
 \resizebox{\textwidth}{!}{%
 \begin{tabular}{l|ccccc}
  \toprule
             & \textbf{4s-ShiftGCN} & \textbf{MS-G3D} & \textbf{VA-CNN} & \textbf{2s-SDGCN} & \textbf{2s-AGCN} \\
  \midrule
    \multirow{10}{4.5em}{\centering Best-10}
    & Put on jacket & Stand up  & Walking towards  & Walking towards & Walking towards \\ 
    & Walking towards & Nod head or bow & Falling down & Put on jacket & Put on jacket\\ 
    & Staggering & Put on jacket & Jump up & Stand up & Stand up\\ 
    & Jump up & Walking towards & Staggering & Nod head or bow & Hopping\\ 
    & Stand up & Arm circles & Hopping & Walking apart & Arm circles \\ 
    & Walking apart & Hopping & Stand up & Hopping & Staggering\\ 
    & Take off jacket & Staggering & Arm swings & Arm circles & Nod head or bow\\ 
    & Hopping & Arm swings & Walking apart & Jump up & Cheer up\\ 
    & Nod head or bow & High five & Cross toe touch & Take off jacket & Drink after cheers\\ 
    & Cross toe touch & Walking apart & Nod head or bow & Sit down & Walking apart\\ 
  \midrule
    \multirow{10}{5em}{\centering Worst-10}
    & Staple book & Writing & Make ok sign & Staple book & Cutting paper\\ 
    & Writing & Staple book & Staple book & Writing & Play magic cube\\ 
    & Make ok sign & Cutting paper & Writing & Cutting paper & Make ok sign\\ 
    & Cutting paper & Make ok sign & Counting money & Yawn & Staple book\\ 
    & Yawn & Counting money & Reading & Counting money & Make victory sign\\ 
    & Make victory sign & Cutting nails & Yawn & Cutting nails & Counting money\\ 
    & Wield knife & Yawn & Make victory sign & Make ok sign & Writing\\ 
    & Counting money & Wield knife & Cutting paper & Make victory sign & Yawn\\ 
    & Reading & Reading & Play with phone or tablet & Reading & Type on keyboard\\ 
    & Cutting nails & Make victory sign & Hit with object & Play magic cube & Cutting on nails\\ 
  \bottomrule
  \end{tabular}
  }
  \captionof{table}{Best-10 and Worst-10 classes for the models trained on NTU120 (Cross Setup) }
  \label{tab:top-bottom-NTU-cset}
\end{table*}

\subsection{Performance on partial sequences}
\label{sec:ntu120-temporal}

Action recognition from partially observed sequences has been an active area of research \cite{temporal1}, \cite{temporal2}, \cite{temporal3}, \cite{temporal4} and has many practical applications in the field of video surveillance and human-computer interaction. The ambiguity induced by partial sequences naturally makes this a challenging problem. To study action recognition in this setting, we benchmark the top-3 models of Table~\ref{tab:ntu120-val} on partially observed skeleton sequences of NTU-120 using the Cross Subject protocol.   The increase in accuracy is on expected lines, i.e. actions are generally better recognized when the extent to which they are accessible increases (see Figure \ref{fig:ntu120-temporal}). Note that 4s-ShiftGCN and MS-G3D outperform the third best performer, VA-CNN, by a noticeable margin. This is likely due to the complementary features which are learnt via the multiple feature stream processing present in 4s-ShiftGCN and MS-G3D models.

\begin{figure*}
\centering
\begin{minipage}[b]{.47\textwidth}
\includegraphics[width=\linewidth]{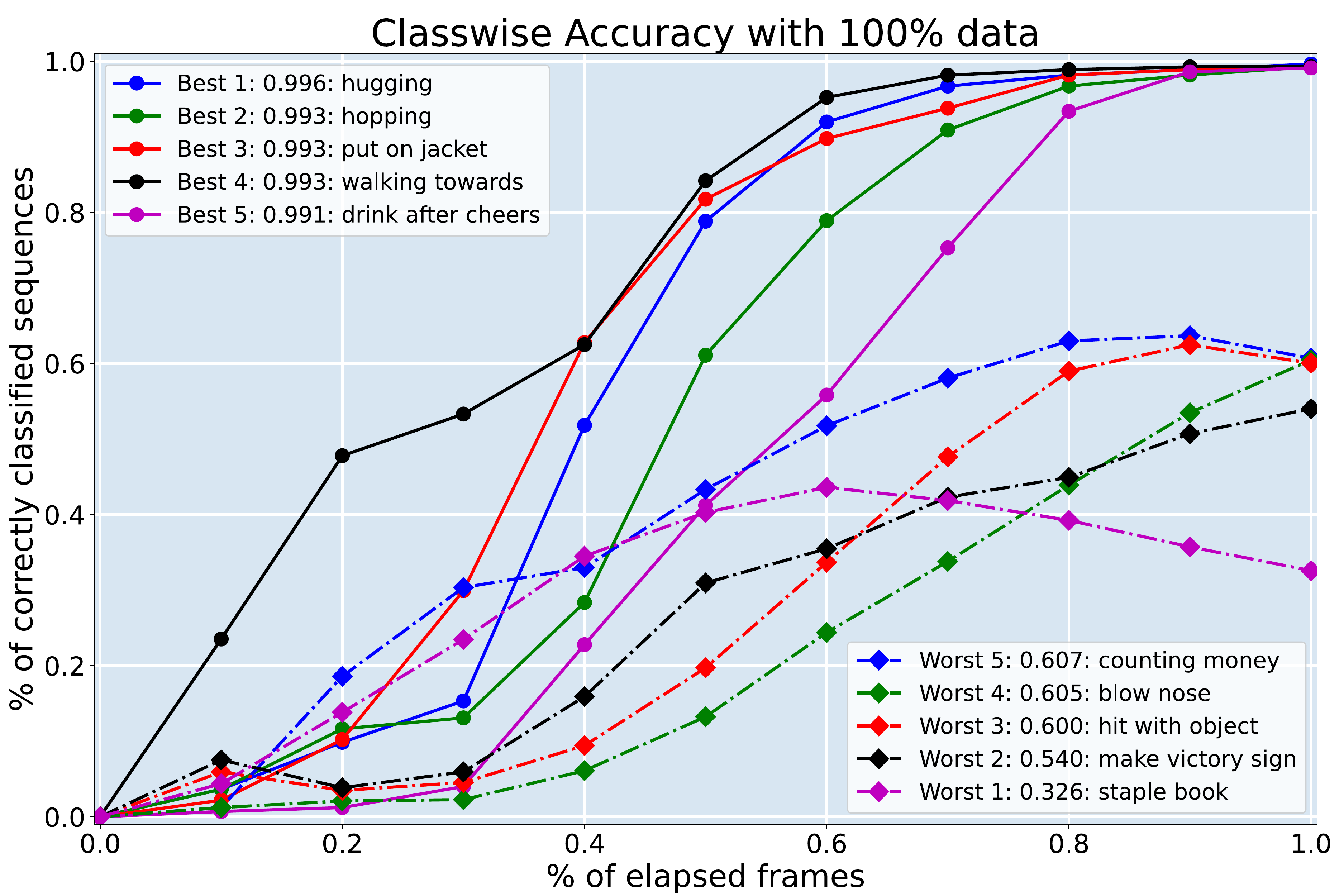}
\caption{Early recognition curves for best-5, worst-5 classes of MS-G3D model on NTU-120 Cross Subject with classwise accuracy as the measure.}\label{temporal_acc}
\end{minipage}
\qquad
\begin{minipage}[b]{.47\textwidth}
\includegraphics[width=\linewidth]{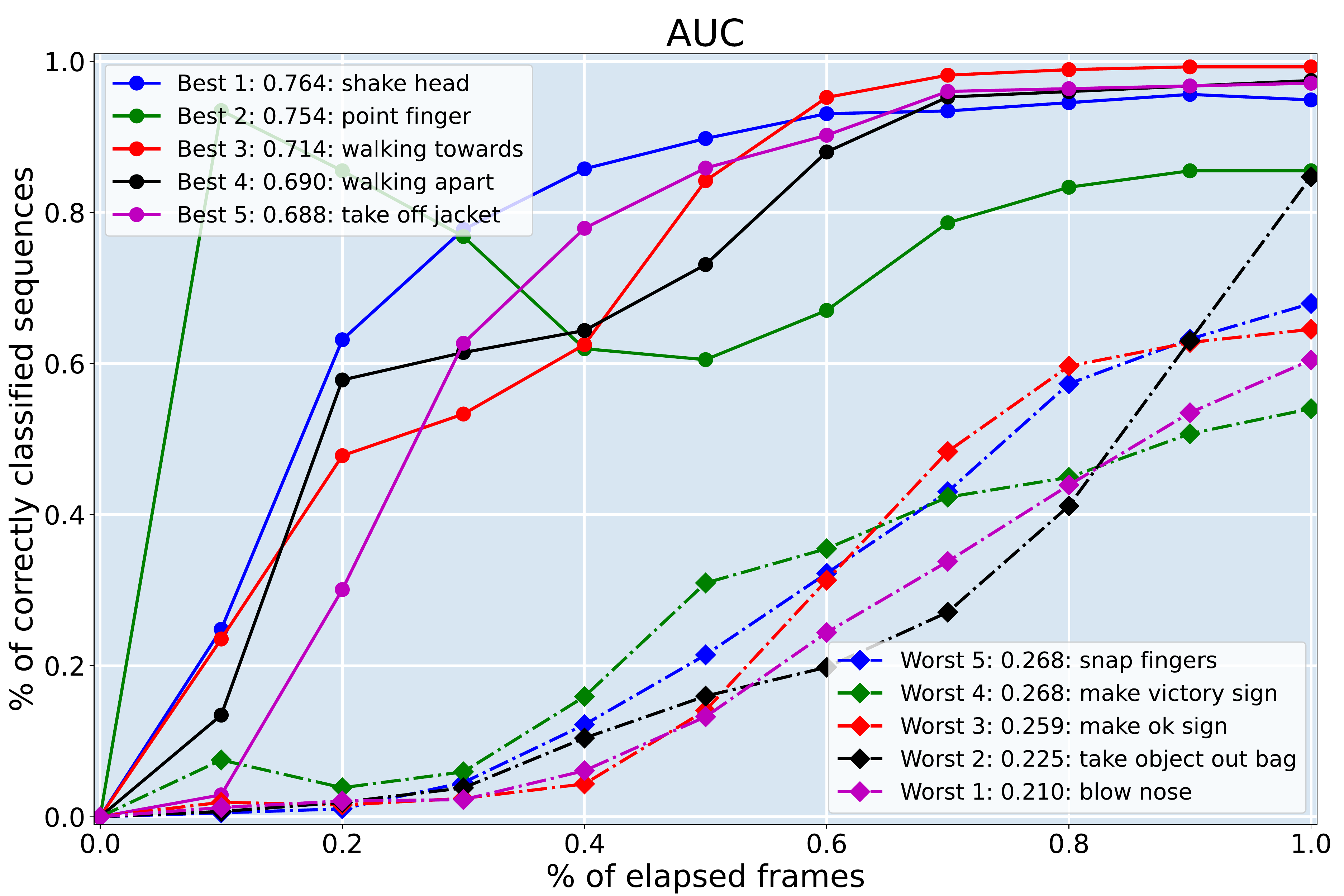}
\caption{Early recognition curves for best-5, worst-5 classes of MS-G3D model on NTU -120 Cross Subject with AUC as the measure.}\label{temporal_auc}
\end{minipage}
\end{figure*}

To explore this at class level, we replicate the plot of Figure~\ref{fig:ntu120-temporal}, but now for the top-1 model (MS-G3D) and for individual action categories. The curves for best-5 and worst-5 classes sorted by \textit{overall} accuracy can be viewed in Figure~\ref{temporal_acc}. The closer an activity's curve is to the top-left corner, the better is its ability to be recognized early. It is evident from the plot that most of the best-5 classes are not confidently recognizable until 50\% of the action sequence is completed. From this viewpoint, it is also interesting to note that some activities (best-3,4) ranked lower than the best-1 but have an earlier onset of recognition. The temporal effects of intra-class confusion on worst ranked classes can also be observed in this plot.

To understand recognition onset trends in a more fine-grained manner, we propose Area-under-curve (AUC) as a better alternative, i.e. the normalized area under the curve with \% of elapsed sequence on x-axis and \% of correctly recognized sequences on the y-axis. The closer a category's AUC to 1, the earlier it can be recognized. Refer to Figure~\ref{temporal_auc} where best-5 and worst-5 activities by AUC can be seen. Multiple interesting trends can be observed. Firstly, most of the best-5 and worst-5 classes are different from overall accuracy plot counterparts from Figure~\ref{temporal_acc}. The AUC-wise top performing activities (e.g. `Shake head', `Walking', and `Take off jacket') contain unique action sequences from the beginning and therefore, are more consistently identified with increasing number of frames.  Also classes like `Walking Towards' and `Walking Apart' which show very minimal intra-class variation and differ only in terms of increasing distance between the subjects are easily differentiable with the increasing frames. The accuracy of `Point finger' decreases in the first half due to barely discernible joint movement in the initial frames. Among the AUC-wise worst classes (`Snap fingers', `Make victory sign', `Make OK sign'), the finger-joint motion is predominant. Since hand joints are not captured via Kinect v2 sensor in NTU-120 dataset, a fundamental bottleneck arises in recognizing these activities regardless of the elapsed time.

\section{Skeleton Action Recognition in the Wild}
\label{sec:skwild}

In the upcoming sections, we describe our experiments using the newly introduced datasets and involving the best performing architectures on NTU-120 dataset discussed in previous section.

\subsection{Skeletics-152}
\label{sec:skeleticsres}

The train-test split in Skeletics-152 is created following  splits originally provided with Kinetics-700. We use the original validation set of Kinetics-700 as our test set. We randomly split the original Kinect-700 training set into training and validation sets in a 85:15 ratio. To address class imbalance, we employ class-frequency based mini-batch resampling and class-based loss weighting.

\begin{table}[!t]
\resizebox{\linewidth}{!}
 {%
  \centering
  \begin{tabular}{c|cc}
 \toprule
            Model & Accuracy & F-1 score\\
 \midrule
MS-G3D (trained from scratch)  &   $56.39$ & $50.80$\\
4s-ShiftGCN (trained from scratch) &  $56.15$ & $50.41$\\
MS-G3D (pretrained on NTU-120 + finetuned) &  $55.75$ & $49.57$\\
 4s-ShiftGCN (pretrained on NTU-120 + finetuned)  &   $\mathbf{57.01}$ & $\mathbf{51.13}$\\
 \bottomrule
 \end{tabular}
 }
\captionof{table}{Results on Skeletics-152 test set with mean accuracy as performance measure.}
\label{tab:Kinetics-results}
\end{table}

\begin{table}[!t]
    \resizebox{\linewidth}{!}
    {
    \begin{tabular}{lcc}
    \toprule
         & MS-G3D &  4s-ShiftGCN  \\
         \midrule
          \multirow{5}{4.5em}{\centering Best-5} & Mountain climber (exercise) & Mountain climber (exercise) \\ 
          & Front raises & Clean and jerk \\ 
          & Jumping Jacks & Front raises \\ 
          & Deadlifting & Lunge\\ 
          &  Lunge & Jumping jacks\\ 
          \midrule
          \multirow{5}{4.5em}{\centering Worst-5} & High fiving & Falling off chair \\ 
          & Cumbia &  Cumbia \\ 
          & Falling off chair & Swinging baseball bat \\ 
          & Hugging (not baby) & Passing American football (not in game)\\ 
          & Combing hair &  Digging\\ 
          \bottomrule
    \end{tabular}
    }
    \captionof{table}{Best-5 and Worst-5 classes of all models trained on Skeletics-152 dataset.}
    \label{tab:top-bottom-skeletics}
\end{table}

We evaluated the best two performers (MS-G3D and 4s-ShiftGCN) from NTU-120 in two training regimes. In one regime, we first extracted VIBE skeletons from RGB videos of NTU-120 and trained the models on these skeletons. This allows us to eliminate the effect of different joint positions in NTU-120 dataset and VIBE pose estimation. The resulting models were ultimately fine-tuned on the Skeletics-152 data. In the second regime, we trained the models from scratch on Skeletics-152 data. We found that 4s-ShiftGCN provides the best performance (Table~\ref{tab:Kinetics-results}). Pre-training on NTU-120 provides a slight benefit compared to training from scratch. Comparing the performance rates in Tables \ref{tab:ntu120-val} and ~\ref{tab:Kinetics-results}, it is evident that skeleton-based action recognition in the wild is significantly more challenging given the inter/intra-category diversity and noise-inducing factors (e.g. occlusion, lighting, uncontrolled background context, interaction with unnecessary objects). In addition, we empirically observed that even the best 3-D pose estimators routinely generate poorly localized joint estimates, impacting performance.

\begin{table}[!t]
\centering
\resizebox{0.9\linewidth}{!}
 {
  \begin{tabular}{c|c|c}
 \toprule
             \diagbox[width=10em]{Base model}{Training set} &
  \makecell{Skeletics-152\\(Complete)} & \makecell{Skeletics-152\\(Skeleton-Mimetics classes)} \\ 
 \midrule
 MS-G3D & $\mathbf{57.37}$  & $49.22$ \\
 4s-ShiftGCN & $56.11$ & $51.10$ \\
\bottomrule
     \end{tabular}
     }
 \captionof{table}{Performance summary in terms of mean accuracy for Skeleton Mimetics dataset as the test set. The 25 joints skeletons for both skeletics-152 (train set) and Skeleton-Mimetics (test set) are exctracted using VIBE. Refer to Sec.~\ref{sec:mimeticsres} for details.}
\label{tab:Mimetics-test-results}
\end{table}

As shown in the Table \ref{tab:top-bottom-skeletics}, the best-5 classes are all exercise based activities which tend to have very low intra class variability. On the other hand, sequences from the worst-5 classes exhibit a lot of diversity and intra-class variability (see Figure \ref{fig:skeletics-mimetics-3}).

\subsection{Skeleton-Mimetics}
\label{sec:mimeticsres}

Following the procedure of Weinzaepfel et. al.~\cite{weinzaepfel2019mimetics}, we use our proposed Skeleton-Mimetics only for evaluation. Similar to Skeletics-152, we use 4s-ShiftGCN and MS-G3D as the base models. Due to the similarity of challenges faced in recording out-of-context skeleton actions and actions in the wild, we choose Skeletics-152 as the training dataset. The final prediction is obtained by considering the maximum softmax score among the 23 skeleton-mimetics classes out of the 152 in Skeletics. We perform another experiment where instead of training on samples from all the 152 classes of Skeletics-152, samples pertaining only to the 23 skeleton mimetics classes are used.  

\begin{table*}[!ht]
 \centering
 \resizebox{\textwidth}{!}{%
 \begin{tabular}{l|c|c|c|c}
  \toprule
             & NTU-60 [MS-G3D] & NTU-120 [MS-G3D] & SKELETICS-152 [4s-ShiftGCN] & SKELETON-MIMETICS [MS-G3D]\\
  \midrule
    \multirow{5}{4.5em}{\centering Best-5}
    & Staggering (99.64\%) &  Hugging (99.64\%) & Mountain climber (exercise) (92.59\%) & Golf driving (93.33\%) \\ 
    & Jump up (99.27\%) & Hopping (99.27\%) & Clean and jerk (89.25\%) & Deadlifting (90.00\%) \\ 
    & Falling down (98.54\%) &  Put on jacket (99.27\%) & Front raises (87.37\%) & Clean and jerk (83.33\%)\\ 
    & Put on jacket (98.53\%)  & Walking towards (99.26\%) & Lunge (87.14\%) & Climbing a rope(78.57\%)\\ 
    & Hopping (98.18\%) & Drink after cheers (99.13\%) & Jumping jacks (87.10\%) & Playing saxophone(75.00\%)\\
  \midrule
  \midrule
    \multirow{5}{4.5em}{\centering Worst-5}
    & Writing (57.41\%) & Staple book (32.57\%) & Falling off chair (11.5\%) & Playing tennis (15.79\%)\\ 
    & Eat meal (71.43\%) & Make victory sign (54.02\%) & Cumbia (12.90\%) & Playing guitar (16.67\%)\\ 
    & Reading (72.43\%) & Hit with object (60.03\%) & Swinging baseball bat (14.29\%) & Reading a book (30.00\%)\\ 
    & Sneeze or cough (77.17\%) & Blow nose (60.45 \%) & Passing American football (not in game) (16.28\%) & Bowling (38.46\%)\\ 
    & Play with phone or tablet (78.75\%) &  Counting money (60.70\%) & Digging (17.39\%) & Catching or throwing baseball (38.46\%)\\
  \bottomrule
  \end{tabular}
  }
  \captionof{table}{List of Best-5 and Worst-5 classes in terms of accuracy for NTU-60, NTU-120, Skeletics-152 and Skeleton-Mimetics datasets. The model associated with the best peformance is in brackets alongside the dataset name.}
 \label{tab:top5bottom5}
\end{table*}

The results shown in Table ~\ref{tab:Mimetics-test-results} depict that for both the base models, training on the complete Skeletics-152 dataset, improves performance as compared to training on the smaller split containing the 23 skeleton mimetics classes. The results suggest that training on a more larger dataset (classes outside the test set) helps the models to learn better generalisable features which boost the test performance, whereas a smaller dataset (classes specific to the test set) tends to learn features which offers limited generalisability.

\subsection{Metaphorics}
\label{sec:MetaphoricsEvaluation}

\begin{table}[!t]
\resizebox{\linewidth}{!}
 {%
  \centering
 \begin{tabular}{l|c|c}
 \toprule
             Method & Training Dataset & Mean Cosine Similarity \\
 \midrule
  4s-ShiftGCN~\cite{cheng2020shiftgcn} & NTU-120 & $0.04$ \\
  MS-G3D~\cite{liu2020disentangling} & NTU-120 & $0.14$\\
  4s-ShiftGCN~\cite{cheng2020shiftgcn} & Skeletics-152 & $0.08$ \\
  MS-G3D~\cite{liu2020disentangling} & Skeletics-152 & $0.10$\\
  \bottomrule
 \end{tabular}
  }
\caption{\label{tab:metaphorics} Benchmarking comparison for Metaphorics dataset (mean cosine similarity). }
\end{table}

Due to the small number of available action segments, we use the Metaphorics dataset only for evaluation. To begin with, we use the previously determined state of the art skeleton action recognition models - MS-G3D and 4s-ShiftGCN to obtain the class predictions for the action segments. Since the label sets used to train the models and the ones from Metaphorics dataset are different, direct comparison is not possible. Therefore, to perform quantitative evaluation, we propose to compare lexical representations of the predicted label and the ground truth label from the Metaphorics dataset. To obtain the lexical representations, we average the \textit{word2vec} embedding vectors of words that constitute the action description. Subsequently, we compute the cosine distance between these representations. These distances are averaged across the dataset. Table~\ref{tab:metaphorics} reports the mean performance for two different sets of models -- one set trained on NTU-120 and another set trained on Skeletics-152. The rather poor performance can be attributed to the fundamentally different nature of training and evaluation settings (see Table~\ref{tab:datasets}). In particular, the abstract, non-contextual, gesture-style actions in Metaphorics are qualitatively distinct from the explicit and contextual actions present in training datasets (NTU-120, Skeletics-152).

To obtain a qualitative perspective, we report top-5 predictions by MS-G3D, 4s-ShiftGCN pre-trained on Skeletics-152 for sample action segments in Figure~\ref{fig:metaphorics-results}. As mentioned before, target phrases in Dumb Charades and interpretative dances are typically enacted indirectly using metaphors. Models trained on other datasets tend to map the skeleton sequence `literally' to action labels. This explains some of the predictions seen in Figure~\ref{fig:metaphorics-results}. For example, the model predictions for actions shown in the first row are related to the ground-truth tags (`fight', `swimming'). The predictions for the other example sequences highlight the shortcomings arising from the literal nature of actions in contextual action datasets as mentioned previously.

\section{Discussion}
\label{sec:qualitative}

In this section, we analyze the salient trends for skeleton action recognition approaches across different datasets. The list of best-5 and worst-5 classes for various scenarios (datasets) can be viewed in Table~\ref{tab:top5bottom5}.

The first two columns correspond to the lab-based indoor datasets - NTU-60 and NTU-120. It is interesting to note that even the worst performing classes of NTU-60 have accuracy in the range 50-70 \% while the counterparts in NTU-120 exist in a much lower range (32-60 \%). One reason is that the introduction of NTU-120 resulted in an increase of action classes with subtle, finger-level movements which impacts performance as mentioned previously (Section~\ref{sec:ntu120-temporal}). Overall, our analysis motivates the need for approaches which can explicitly focus on boosting the performance for classes ranked lowest. Another concurrent requirement arising from our analysis is for skeleton representations which provide finger-level joint information.

We have already seen that average performance in the wild is relatively lower compared to lab-based settings (Tables~\ref{tab:ntu120-val},~\ref{tab:Kinetics-results}). The results from Table~\ref{tab:top5bottom5} for Skeletics-152 reflect this trend as well. Actions in the wild exhibit large intra-class variability which affects even the best-5 classes (cf. best-5 of NTU-120). Actions belonging to the worst-5 classes in Skeletics-152 and Skeleton-Mimetics are characterized either by high intra-class variability or by containing subtle, finger-dominant motions which cannot be captured by existing skeleton representations. Additionally, action sequences in NTU-120 are somewhat choreographed, having a defined starting pose and ending pose, but this is absent in Skeletics.

In terms of base architectures, MS-G3D provides the best performance across various datasets except for Skeletics-152, where 4s-ShiftGCN is the best performer.  A pictorial illustration of performance trends in the top-2 models for selected action classes from Skeletics-152 and Skeleton-Mimetics can be viewed in Figure~\ref{fig:skeletics-mimetics-3}. Interestingly, even for the classes with lowest performance (worst-3), the correct prediction for Skeletics is often in the list of top-5 model predictions. This is similar to the trend already observed for NTU-120 (Section~\ref{sec:ntu120}). 

Such class-level insights cannot be deduced for the proposed Metaphorics dataset since its label set is open-ended, i.e. does not contain a fixed set of action categories. We believe this setting represents an open frontier for skeleton action recognition. The performance on skeleton version of Metaphorics provides an opportunity to study the generalization capabilities and limitations of existing approaches which are typically optimized for non-interactive, category-based, closed-world recognition paradigms.

\section{Conclusion}

In this paper, we have examined multiple existing and upcoming frontiers in the landscape of skeleton-based human action recognition. As an important facet of establishing new frontiers for skeleton action understanding in the wild, we curate and introduce three new datasets -- Skeletics-152, Skeleton-Mimetics and Metaphorics. Our experiments and benchmarking reveal the capabilities and shortcomings of state-of-the-art recognition models. In addition, the results also highlight the bias induced by processing components (e.g. RGB 3-D pose estimation) and the task paradigm (classification). We hope these findings and the newly introduced datasets will spur the design of better models for `in the wild'  actions, both contextual and non-contextual  -- see the work of Moon et. al.~\cite{moon2020integralaction} as a preliminary representative example.

As mentioned earlier, our findings can be interactively explored at \url{https://skeleton.iiit.ac.in/}. The website features an interactive data analytics dashboard, code and pre-trained models for top-performing skeleton action recognition models and new skeleton action datasets (Skeletics-152, Skeleton-Mimetics, Metaphorics) introduced by us for additional exploration and benefit of the community.

In our current work, we have not examined approaches which map skeleton actions to lexical phrase representations (cf. class labels)~\cite{hahn2019action2vec,jasani2019skeleton} in detail. We intend to study this promising frontier in the future.

\noindent \textbf{Acknowledgements:} We wish to thank the anonymous reviewers for their detailed and constructive feedback. We also wish to thank Kalyan Adithya and Sai Shashank Kalakonda for their efforts in creating the project page. This work is partly supported by MeitY, Government of India.

\bibliographystyle{spmpsci} 
\bibliography{main}

\end{document}